\documentclass[pdflatex,sn-mathphys-num]{sn-jnl}

\usepackage{graphicx}%
\usepackage{multirow}%
\usepackage{amsmath,amssymb,amsfonts}%
\usepackage{amsthm}%
\usepackage[scr=rsfs]{mathalpha}%
\usepackage[title]{appendix}%
\usepackage{xcolor}%
\usepackage{textcomp}%
\usepackage{manyfoot}%
\usepackage{booktabs}%
\usepackage{algorithm}%
\usepackage{algorithmicx}%
\usepackage{algpseudocode}%
\usepackage{listings}%
\usepackage{makecell}%
\usepackage{diagbox}%
\usepackage[T1]{fontenc}%
\usepackage{graphicx}

\theoremstyle{thmstyleone}%
\theoremstyle{thmstyletwo}%

\theoremstyle{thmstylethree}%

\begin{document}

\title[Article Title]{FSPGD: Rethinking Black-box Attacks on Semantic Segmentation}


\author[1,2]{\fnm{Eun-Sol} \sur{Park}}\email{espark\_82@korea.ac.kr}

\author[1,2]{\fnm{Miso} \sur{Park}}\email{miso419@korea.ac.kr}

\author*[1,2]{\fnm{Yong-Goo} \sur{Shin}}\email{ygshin92@korea.ac.kr}

\affil[1]{\orgdiv{Department of Electronics and Information Engineering}, \orgname{Korea University}, \orgaddress{\city{Sejong}, \postcode{30019}, \country{Republic of Korea}}}

\affil[2]{\orgdiv{Division of Smart Energy Convergence Engineering}, \orgname{Korea University}, \orgaddress{\city{Sejong}, \postcode{30019}, \country{Republic of Korea}}}

\abstract{Black-box adversarial attacks on semantic segmentation remain a challenging problem, particularly in the black-box transfer attack setting where perturbations crafted on a surrogate model are expected to mislead unseen target models. Existing methods typically operate only on output logits and thus fail to account for the spatial structure and class-wise feature relationships that are crucial for dense prediction. To address this limitation, we propose Feature Similarity Projected Gradient Descent (FSPGD), a feature-space black-box attack that explicitly disrupts intermediate representations. FSPGD employs a dual loss design: an external loss that enforces discrepancy between clean and adversarial features to weaken cross-model alignment, and an internal loss that reduces feature consistency among spatially separated instances of the same class. Comprehensive experiments on Pascal VOC 2012 and Cityscapes across both CNN-based and Transformer-based backbones demonstrate that FSPGD achieves state-of-the-art transferability, consistently outperforming conventional logit-level methods as well as recent segmentation-specific baselines such as SegPGD, CosPGD, and RP-PGD. Moreover, adversarial training with FSPGD examples enhances robustness against unseen attacks across multiple architectures, further validating the effectiveness of our design. These findings establish FSPGD as a principled and practical framework for advancing black-box adversarial attacks in semantic segmentation. Code is available at \href{https://github.com/KU-AIVS/FSPGD}{https://github.com/KU-AIVS/FSPGD}.}


\maketitle

\section{Introduction}\label{sec:intro}

Convolutional neural networks (CNNs), have demonstrated remarkable performance across a wide range of computer vision tasks over the past decade. In image classification, models such as VGG~\cite{simonyan2014very}, ResNet~\cite{he2016deep}, DenseNet~\cite{huang2017densely}, and Inception~\cite{szegedy2015going} have successively advanced the state of the art. Beyond classification, CNN-based methods have also transformed dense prediction tasks such as semantic segmentation~\cite{long2015fully, chen2017rethinking, chen2018encoder, zhao2017pyramid}, where pixel-level understanding of scenes is required, and image synthesis~\cite{goodfellow2020generative, park2024rethinking, rombach2022high, sagong2022conditional}, where CNNs and diffusion-based models produce realistic and diverse content. These advances have enabled practical deployment of computer vision models in critical domains including autonomous driving, medical imaging, surveillance, and human–computer interaction. However, despite their widespread success, CNNs remain highly vulnerable to adversarial perturbations, which are carefully designed but imperceptible noise patterns that can significantly degrade model performance. This vulnerability is particularly concerning in safety-critical applications such as self-driving vehicles~\cite{eykholt2018robust} and facial verification systems~\cite{sharif2016accessorize}, where even small prediction errors may lead to catastrophic outcomes.

Adversarial attacks can be broadly categorized into two groups: white-box and black-box attacks~\cite{dong2018boosting, lin2019nesterov, xie2019improving}. In the white-box setting, the attacker is assumed to have complete access to the architecture, parameters, and gradients of the target model, allowing highly effective perturbations to be crafted directly by backpropagation. Methods such as FGSM~\cite{goodfellow2014explaining}, PGD~\cite{mkadry2017towards}, MI-FGSM~\cite{dong2018boosting}, and NI-FGSM~\cite{lin2019nesterov} exemplify this paradigm and have become standard benchmarks in adversarial robustness research. While white-box attacks typically achieve strong attack performance, they are often unrealistic in practice because detailed knowledge of deployed models is rarely available. In contrast, black-box attacks assume no access to model parameters or gradients, which better reflects real-world deployment scenarios. Within the black-box category, query-based attacks~\cite{chen2017zoo, ilyas2018black} iteratively query the target model to estimate gradients or decision boundaries, while transfer-based black-box attacks exploit perturbations generated on a surrogate model in the hope that they transfer to unseen targets. Among these, transfer-based attacks are particularly important since they require no interaction with the target model and can succeed when only a surrogate trained on similar data is available. However, their effectiveness crucially depends on transferability, that is, the ability of adversarial examples generated on one model to remain effective against others with different architectures and training procedures.

For image classification, transfer-based black-box attacks have been extensively studied, and numerous strategies have been proposed to improve transferability. Input transformation methods such as DI-FGSM~\cite{xie2019improving}, TI-FGSM~\cite{dong2019evading}, and Admix~\cite{wang2021admix} diversify perturbations by altering the input space. Optimization strategies including MI-FGSM~\cite{dong2018boosting}, NI-FGSM~\cite{lin2019nesterov}, and momentum-based variants enhance gradient stability and improve cross-model effectiveness. Feature-level approaches~\cite{huang2019enhancing, wang2021feature, zhang2022improving, li2024improving, weng2023boosting} leverage intermediate representations to generate more transferable perturbations by aligning or disrupting features across models. These advances have established strong baselines in classification, but directly applying them to semantic segmentation is nontrivial. Unlike classification, where predictions involve a single object label, segmentation produces dense pixel-wise predictions involving multiple object classes, spatially separated instances of the same class, and long-range contextual relationships. The inherent complexity of segmentation makes it difficult for adversarial examples crafted on one model to generalize to others, as even small changes in feature extraction or context modeling can significantly affect transferability.

Recent works have attempted to address these challenges by designing attacks tailored for segmentation. SegPGD~\cite{gu2022segpgd} extended PGD to segmentation and demonstrated that iterative attacks remain effective in dense prediction. TranSegPGD~\cite{jia2023transegpgd} improved upon SegPGD by introducing a two-stage procedure to enhance transferability. CosPGD~\cite{agnihotri2024cospgd} proposed a cosine-similarity-based formulation, while other studies have explored ensemble-based perturbations~\cite{chen2023adaptive} or theoretical analyses of logit-space attacks~\cite{chen2023rethinking}. Despite these contributions, a fundamental limitation persists: most segmentation attacks remain logit-level methods that directly perturb output predictions, similar to classification-based attacks. Such approaches overlook the fact that semantic segmentation involves complex contextual information and intra-class relationships, particularly when multiple objects of the same class appear in spatially separated regions. As a result, the transferability of existing methods remains limited, as confirmed by our empirical analysis and by recent critiques in the literature.

To overcome these limitations, we propose a new black-box attack method termed \textit{Feature Similarity Projected Gradient Descent (FSPGD)}. The central idea of FSPGD is to move beyond logit-level perturbations and operate directly in the intermediate feature space, where both local and contextual information are encoded. FSPGD introduces a dual loss function to explicitly disrupt representation similarity. The external similarity loss enforces discrepancy between clean and adversarial features, thereby weakening cross-model feature alignment. The internal similarity loss reduces feature consistency among spatially separated instances of the same class, increasing perturbation diversity and directly addressing the segmentation-specific challenge of intra-class similarity. 

Extensive experiments on Pascal VOC 2012 and Cityscapes validate the effectiveness of FSPGD. Across both CNN-based and Transformer-based models, FSPGD consistently outperforms conventional logit-level attacks and recent segmentation-specific baselines including SegPGD, CosPGD, and RP-PGD. Furthermore, adversarial training with FSPGD examples improves robustness against unseen attacks, underscoring the utility of our method for both attack and defense. Beyond empirical results, ablation studies clarify the complementary roles of the proposed loss components and confirm the robustness of FSPGD across hyperparameter choices and feature extraction layers. These contributions highlight FSPGD as a principled and effective framework for advancing black-box adversarial attacks in semantic segmentation.

The main contributions of this work are as follows:
\begin{itemize}
    \item We propose FSPGD, a feature-space black-box attack that integrates external and internal similarity losses to jointly disrupt cross-model alignment and intra-class consistency.
    \item We provide extensive experiments on multiple datasets and model architectures, including CNNs and Transformers, and conduct comprehensive ablation studies to evaluate the roles of loss terms, parameter settings, and feature extraction layers.
    \item We demonstrate that adversarial training with FSPGD-generated examples enhances robustness against a wide range of unseen attacks, confirming the broader applicability of the proposed framework.
\end{itemize} 

\section{Preliminaries}\label{sec:preliminaries}

\subsection{Problem Setting}
Let $x \in [0,1]^d$ denote a clean input image and $y\in\mathbb{R}^{H \times W \times C}$ the corresponding ground-truth segmentation label map, where $H \times W$ is the spatial resolution and $C$ the number of semantic classes. A segmentation network $F(\cdot;\theta)$ produces per-pixel logits $Z = F(x;\theta)\in\mathbb{R}^{H \times W \times C}$, which are converted into probabilities $P=\mathrm{softmax}(Z)$, and the final prediction $\hat{y}(u)=\arg\max_c P(u,c)$ for each pixel $u$. 

In this work we focus on the \emph{transfer-based black-box} threat model. The adversary is assumed to have no access to the parameters or gradients of the target model $F_t(\cdot;\theta_t)$, and is restricted to perturbations generated on a surrogate model $F_s(\cdot;\theta_s)$. Adversarial examples are then transferred to target models to evaluate their effectiveness. Perturbations must remain imperceptible, which we formalize as belonging to an $\ell_p$ ball of radius $\epsilon$ around the clean image and constrained to the valid image domain:
\[
\mathcal{B}_p(x,\epsilon)=\{x'\in[0,1]^d:\ \|x'-x\|_{p}\le\epsilon\}.
\]
Following common practice in adversarial segmentation~\cite{gu2022segpgd, jia2023transegpgd, agnihotri2024cospgd, chen2023adaptive, chen2023rethinking, xie2017adversarial}, we adopt the $\ell_\infty$ norm, which constrains each pixel perturbation to at most $\epsilon$. The attacker seeks an adversarial example $x^{adv}$ that maximizes a surrogate loss $\mathcal{L}$ under these constraints:
\begin{equation}
\label{eq1}
    x^{adv} = \underset{\|x'-x\|_{p}\le \epsilon,\ x'\in\mathcal{X}}{\mathrm{argmax}}\ \mathcal{L}(x',y;\theta_s) , \qquad \mathcal{X}=[0,1]^d.
\end{equation}
Unless otherwise specified, $\mathcal{L}$ is defined as the mean per-pixel cross-entropy between the prediction and the ground-truth label map. 

A well-known baseline is the Fast Gradient Sign Method (FGSM)~\cite{goodfellow2014explaining}, which performs a single-step update for $p=\infty$ by moving in the sign of the gradient:
\begin{equation}
\label{eq2}
    x^{adv} = \mathrm{clip}_{\mathcal{X}}\!\big(x + \epsilon\cdot \mathrm{sign}\!\big(\nabla_x \mathcal{L}(x,y;\theta_s)\big)\big).
\end{equation}
Although efficient, this approach produces relatively weak perturbations, particularly for dense prediction tasks such as segmentation.

\subsection{Attack Optimizers and Baselines}
To obtain stronger attacks, iterative optimization is employed. Projected Gradient Descent (PGD)~\cite{mkadry2017towards} is one of the most widely adopted iterative schemes and forms the foundation of many segmentation attacks~\cite{gu2022segpgd, agnihotri2024cospgd, xie2017adversarial, arnab2018robustness, nokabadi2024trackpgd, savostianova2024low, waghela2024enhancing, huang2023t}. Starting from a randomly perturbed initialization $x^{adv}_0 = \mathrm{clip}_{\mathcal{X}}(x + \eta)$ with $\eta \sim \mathcal{U}([-\epsilon,\epsilon]^d)$, the attack is iteratively updated as
\begin{equation}
\label{eq3}
\begin{aligned}
x^{tmp}_t &= x^{adv}_{t-1} + \alpha \cdot \mathrm{sign}\!\big(\nabla_x \mathcal{L}(x^{adv}_{t-1},y;\theta_s)\big), \\
x^{adv}_t &= \Pi_{\mathcal{B}_p(x,\epsilon)\cap \mathcal{X}}(x^{tmp}_t),
\end{aligned}
\end{equation}
where $\alpha$ is the step size and $\Pi$ denotes projection onto the feasible set $\mathcal{B}_p(x,\epsilon)\cap \mathcal{X}$. PGD has been shown to generate stronger adversarial examples than single-step methods and serves as a standard baseline.

Numerous extensions of PGD have been proposed to enhance transferability in classification tasks. Momentum and Nesterov variants~\cite{dong2018boosting, lin2019nesterov} stabilize gradient directions across iterations. Input transformations such as random resizing and cropping~\cite{xie2019improving} or admixture strategies~\cite{wang2021admix} diversify perturbations. Feature-level methods~\cite{huang2019enhancing, wang2021feature, zhang2022improving} explicitly optimize intermediate representation alignment. These approaches are largely orthogonal to our design and could be integrated into the proposed method, but for clarity of analysis we retain the PGD backbone as our optimization core.

For segmentation, dedicated baselines include SegPGD~\cite{gu2022segpgd}, which adapts PGD to dense prediction, CosPGD~\cite{agnihotri2024cospgd}, which exploits cosine similarity in the logit space, and other recent methods such as TranSegPGD~\cite{jia2023transegpgd}, adaptive ensemble attacks~\cite{chen2023adaptive}, and theoretical reformulations~\cite{chen2023rethinking, he2023transferable}. Despite their contributions, most of these methods operate at the output logit level and thus fail to address the deeper representational similarities that hinder transferability. This motivates our proposal to move beyond logits and explicitly perturb intermediate-layer features.

\begin{figure*}
\centering
\includegraphics[width=1\linewidth]{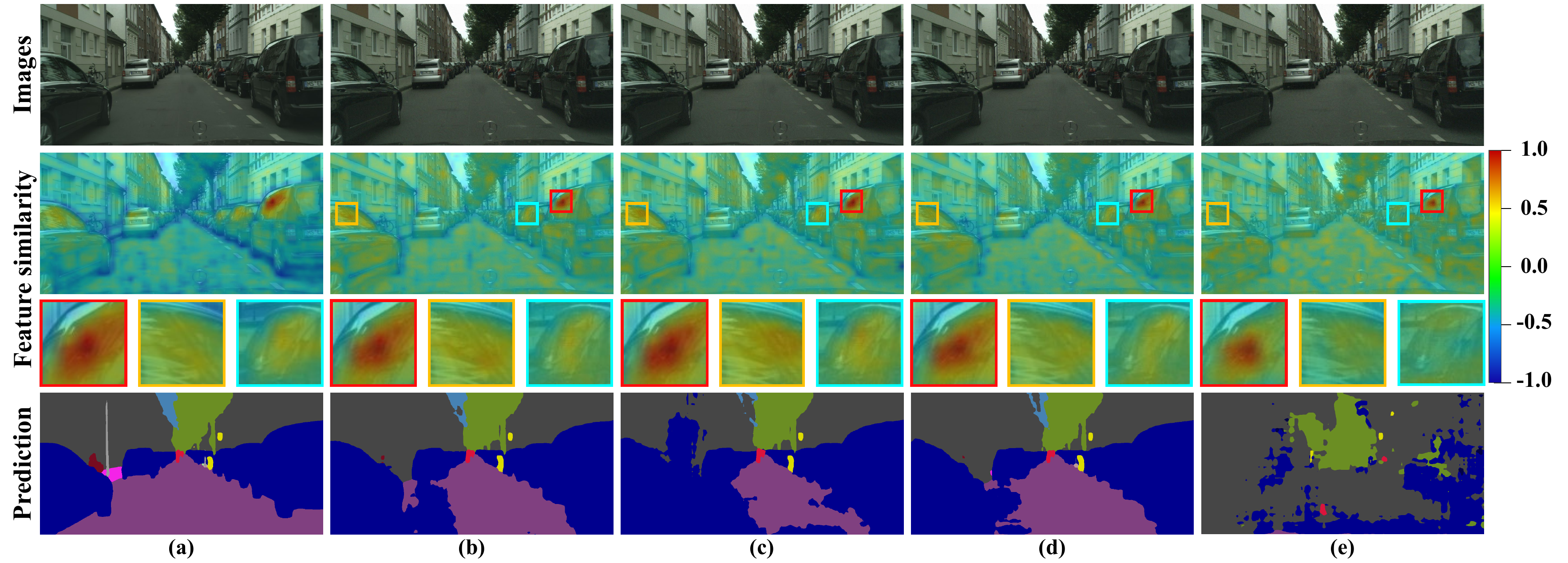}
\caption{Visualization of the feature similarity. We show a feature similarity map using the features of the car area (red box) as the reference feature. In conventional methods, high feature similarity is observed with other cars (yellow and blue boxes), whereas in the proposed method, feature similarity is notably reduced. (a) Clean image, (b) PGD~\cite{mkadry2017towards}, (c) SegPGD~\cite{gu2022segpgd}, (d) CosPGD~\cite{agnihotri2024cospgd}, and (e) FSPGD (Ours).}
\label{fig:fig1}
\end{figure*}

\section{Motivation and Empirical Observation}\label{sec:motivation}

Adversarial transferability in semantic segmentation is strongly shaped by how models encode intermediate representations. Unlike image classification, which predicts a single label at the image level, semantic segmentation produces dense outputs that must capture spatial relationships, contextual dependencies, and multiple object instances. These characteristics create unique challenges for transfer-based black-box attacks that are not adequately addressed by perturbations applied only to output logits.

Our analysis reveals a prominent phenomenon that limits transferability: \emph{intra-class over-consistency} in the feature space. Segmentation networks tend to embed spatially separated instances of the same class into highly similar feature vectors. In Fig.~\ref{fig:fig1}, when the feature vector taken from one car region (red box) is used as the reference, other car regions (yellow and blue boxes) exhibit strong similarity despite being at different spatial locations. This over-consistency implies that perturbations generated around one instance are redundantly replicated on other instances of the same class, reducing the diversity of the adversarial signal and making it less likely to generalize to unseen target models. Conventional logit-level attacks optimize per-pixel predictions but do not explicitly enforce diversity in the representation space, which leaves this limitation largely unaddressed.

\begin{figure*}
\centering
\includegraphics[width=1\linewidth]{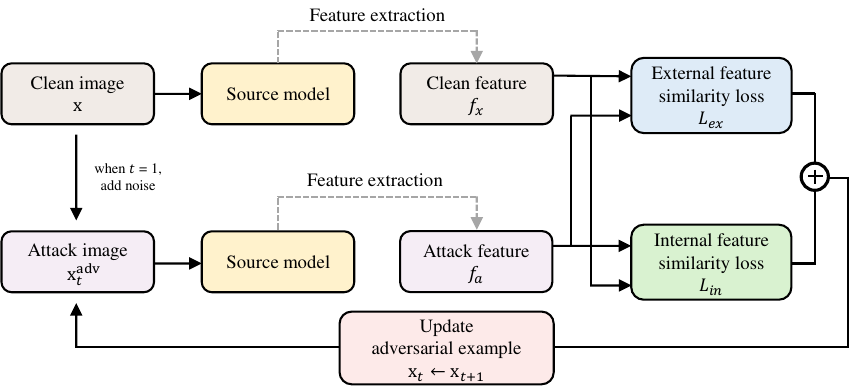}
\caption{Overall framework of FSPGD. FSPGD employs a loss function with two components: external and internal feature similarity loss. The external feature similarity loss measures similarity between intermediate-level features of the clean image and adversarial example, whereas the internal feature similarity loss compares intermediate-level feature similarity among similar objects within adversarial example.}
\label{fig:fig2}
\end{figure*}

The proposed design follows directly from this observation. We target intermediate features and introduce a dual-loss objective that acts on two complementary aspects of the representation. First, an \emph{internal loss} encourages dissimilarity among features of spatially separated instances that the surrogate considers similar in the clean image, thereby reducing intra-class over-consistency and increasing the diversity of perturbations within an image. Second, an \emph{external loss} promotes representational drift between clean and adversarial features at corresponding spatial locations, ensuring that the perturbed features do not collapse back toward their clean counterparts during optimization. These objectives are combined within a projected gradient framework and scheduled over iterations so that early updates emphasize intra-class diversification and later updates strengthen overall feature displacement.

\section{Proposed Method}
\label{sec:proposed}

The proposed Feature Similarity Projected Gradient Descent (FSPGD) is an intermediate–feature space attack designed to overcome the representational limitation identified in Sec.~\ref{sec:motivation}, namely the tendency of semantic segmentation networks to produce highly redundant features for spatially separated instances of the same class. Our method explicitly constructs a dual-objective function in feature space that simultaneously (i) drives adversarial features away from their clean counterparts and (ii) diversifies features across intra-class instances. This dual design is optimized with a PGD backbone under the $\ell_\infty$ threat model introduced in Sec.~\ref{sec:preliminaries}. The overall pipeline is depicted in Fig.~\ref{fig:fig2}.

\subsection{Feature-space Formulation}
\label{sec:method_formulation}
Let $f\in\mathbb{R}^{c\times N}$ denote an intermediate feature map with $c$ channels and $N$ spatial locations extracted from a fixed layer of the surrogate model $F_s(\cdot;\theta_s)$. For a clean input $x$, the extracted feature map is denoted as $f_x\in\mathbb{R}^{c\times N}$, and for the adversarial estimate $x^{adv}_t$ at PGD iteration $t$, the feature map is $f_a\in\mathbb{R}^{c\times N}$. A successful transfer-based black-box attack should enforce two conditions: (i) $f_a$ should become sufficiently dissimilar from $f_x$, so that target models do not process the perturbed input as if it were clean, and (ii) $f_a$ should avoid collapsing multiple instances of the same class into redundant, overly consistent feature vectors. These two requirements are formalized as the external similarity loss $L_{ex}$ and the internal similarity loss $L_{in}$, respectively.

The first component encourages representational drift between $f_x$ and $f_a$ at the same pixel location. Specifically, we penalize cosine similarity:
\begin{equation}
\label{eq4}
    L_{ex} \;=\; -\frac{1}{N}\sum_{i=1}^{N}
    \frac{f_x(i)}{\|f_x(i)\|_2}^{\!\top}\,\frac{f_a(i)}{\|f_a(i)\|_2},
\end{equation}
where $i$ indexes spatial locations. By maximizing $L_{ex}$, we reduce the alignment between clean and adversarial features, thereby countering the tendency of adversarial examples to remain too close to clean representations even when they mislead the surrogate. This mechanism directly addresses the cross-model alignment effect: if $f_a$ drifts significantly from $f_x$, then target models trained on the same data distribution are less likely to interpret $x^{adv}$ as a clean image.

The second component disrupts redundancy among features corresponding to spatially separated instances of the same class. We begin by computing the adversarial-feature Gram matrix $S\in\mathbb{R}^{N\times N}$:
\begin{equation}
\label{eq5}
S(p,q)\;=\;
\begin{cases}
\displaystyle \frac{f_a(p)}{\|f_a(p)\|_2}^{\!\top}\,\frac{f_a(q)}{\|f_a(q)\|_2}, & p\neq q,\\[6pt]
0, & p=q,
\end{cases}
\end{equation}
which encodes pairwise cosine similarity among all spatial locations. To avoid applying this uniformly to all pixel pairs, we leverage information from the clean features $f_x$ to identify pairs that are highly similar under normal conditions. Specifically, we form
\begin{equation}
\label{eq6}
M(p,q)\;=\;
\begin{cases}
\displaystyle \frac{f_x(p)}{\|f_x(p)\|_2}^{\!\top}\,\frac{f_x(q)}{\|f_x(q)\|_2}, & p\neq q,\\[6pt]
0, & p=q,
\end{cases}
\end{equation}
where high values of $M(p,q)$ indicate that pixels $p$ and $q$ belong to objects that the surrogate considers semantically similar. We then binarize this similarity with a threshold $\tau$:
\begin{equation}
\label{eq7}
M_B(p,q)\;=\;
\begin{cases}
1, & \text{if } M(p,q) > \tau,\\[4pt]
0, & \text{otherwise}.
\end{cases}
\end{equation}
Here $M_B$ selects only those pixel pairs that are strongly related in the clean image. Let $K=\sum_{p=1}^{N}\sum_{q=1}^{N} M_B(p,q)$ denote the number of selected pairs. The internal loss is defined as
\begin{equation}
\label{eq8}
    L_{in} \;=\; -\frac{1}{2K}\sum_{p=1}^{N}\sum_{q=1}^{N} M_B(p,q)\,S(p,q),
\end{equation}
where the factor $1/2$ prevents double-counting of symmetric pairs. Maximizing $L_{in}$ forces adversarial features to reduce similarity among selected pairs, thus diversifying intra-class feature representations and directly mitigating the redundancy that weakens transferability in conventional methods.

\subsection{Combined Objective and Optimization}
\label{sec:method_optimization}
The overall objective is a convex combination of the two losses:
\begin{equation}
\label{eq9}
L \;=\; \lambda_t\,L_{ex} \;+\; (1-\lambda_t)\,L_{in}, \qquad \lambda_t\in[0,1].
\end{equation}
We employ a simple yet effective linear schedule $\lambda_t=t/T$ over $t=0,\dots,T-1$. This design emphasizes $L_{in}$ during early iterations, when diversifying intra-class features is most beneficial, and gradually shifts focus toward $L_{ex}$ in later iterations to maximize representational drift. Our ablation studies in Sec.~\ref{sec4.5:Ablation} confirm that this staged combination is significantly more effective than using either term alone or combining them with fixed weights. By explicitly targeting both external alignment and intra-class redundancy through the dual loss and by employing a schedule that exploits their complementarity, FSPGD generates perturbations that transfer robustly across architectures while remaining computationally efficient.

\begin{algorithm}[t]
    \caption{Algorithm of FSPGD}
    \textbf{Input:} Clean image $\mathrm{x}$; clean image feature map $\textit{f}_{x}(\cdot)$; adversarial example feature map $\textit{f}_{a}(\cdot)$; attack iterations $T$; the maximum magnitude of adversarial perturbation $\epsilon$; step size $\alpha$; $\phi^\epsilon(\cdot)$ is a function that clips output into the range $[\mathrm{x}-\epsilon, \mathrm{x}+\epsilon]$; $\mathcal{U}(-\epsilon, \epsilon)$ is a function that initializes random noise into the range $[-\epsilon, \epsilon]$.
    \\
    \textbf{Output:} The adversarial example $\mathrm{x}_{T}^{adv}$ 
    \begin{algorithmic}[1]
    \State \textbf{Initialize} $\mathrm{x}^{adv}_{0}=\mathrm{x}+\mathcal{U}(-\epsilon,\epsilon)$
        \For {$t \leftarrow 0 \ to \ T-1$}
            \State $\lambda_{t} \leftarrow t / T $
            \State Calculate $L_{ex}$ by using Eq.(4)
            \State Calculate $L_{in}$ by using Eq.(8)
            \State $L = \lambda_{t} L_{ex} + (1 - \lambda_{t}) L_{in}$
            \State Calculate the gradient of $L$ with respect to $\textrm{x}^{adv}_{t}$
            \State Update $\textrm{x}^{adv}_{t+1}$
            $$ \textrm{x}^{adv}_{t+1} \leftarrow \textrm{x}^{adv}_{t} + \alpha \cdot sign(\triangledown_{\textrm{x}^{adv}_{t}}L)$$
            \State Clamp on $\epsilon$-ball of clean image 
            $$\textrm{x}^{adv}_{t+1} \leftarrow \phi^\epsilon(\textrm{x}^{adv}_{t+1}) $$
        \EndFor
    \end{algorithmic}
\label{alg1}
\end{algorithm}

\begin{table}[htbp]
\caption{Attack performance comparison on Pascal VOC 2012 in terms of mIoU. Lower mIoU means better performance and bold numbers denote the best mIoU values for each experimental setup}
\setlength{\tabcolsep}{5pt}
\begin{tabular}{c | c | c  c  c  c}
\toprule
& & \multicolumn{4}{c}{Target Models (mIoU$\downarrow$)} \\
\hline
\multirow{2}*{Source Models} & Attack Method & Source Model & PSPRes101 & DV3Res101 & FCNVGG16\\
& Clean Images & 80.22/80.18 & 78.39 & 82.88 & 59.80 \\
\hline
\multirow{9}*{PSPRes50} & PGD~\cite{mkadry2017towards} & 7.72 & 54.73& 59.41     & 45.70  \\
& SegPGD~\cite{gu2022segpgd}& 5.41  & 54.10  & 58.95    & 45.43  \\
& CosPGD~\cite{agnihotri2024cospgd} &1.84   & 56.63 & 64.37    & 45.99   \\
& RP-PGD~\cite{zhang2025rp} & \textbf{0.38} &  63.80 & 69.96 & 46.85 \\
& DAG~\cite{xie2017adversarial}& 65.82  & 62.67 & 66.22& 38.91         \\
& NI-FGSM~\cite{lin2019nesterov}&  7.71 & 33.49   & 38.52   & 32.94  \\
& DI-FGSM~\cite{xie2019improving} & 6.41& 32.00 & 35.25   & 37.34    \\
& TI-FGSM~\cite{dong2019evading} & 18.28 & 64.50 & 69.60& 36.80  \\
& FSPGD (Ours) &  3.39 & \textbf{22.24} & \textbf{16.84} & \textbf{19.75} \\
\hline
\multirow{9}*{DV3Res50} & PGD~\cite{mkadry2017towards}& 9.74 & 52.96 &  56.35 & 46.39 \\
& SegPGD~\cite{gu2022segpgd} & 7.26 &52.05 & 56.50 & 46.23 \\
& CosPGD~\cite{agnihotri2024cospgd} &1.67 & 56.82 &  61.36 & 45.94 \\
& RP-PGD~\cite{zhang2025rp} & \textbf{0.29} & 63.40 & 68.67 & 46.41 \\
& DAG~\cite{xie2017adversarial} & 66.78 & 62.12 & 66.84 & 38.77 \\
& NI-FGSM~\cite{lin2019nesterov}& 9.89 & 33.86 &36.85 & 34.92 \\
& DI-FGSM~\cite{xie2019improving} & 7.35 &31.93 & 32.93 &38.30 \\
& TI-FGSM~\cite{dong2019evading}  & 19.34 &64.99 & 69.80 & 37.65 \\
& FSPGD(Ours) & 3.44 & \textbf{21.89} & \textbf{16.57} & \textbf{19.36} \\
\hline
\hline
\multirow{9}*{PSPRes101}& PGD~\cite{mkadry2017towards} &10.13 & 55.39 &  55.39 &  47.25 \\
& SegPGD~\cite{gu2022segpgd} & 7.31 & 53.56 & 54.03  & 46.26 \\
& CosPGD~\cite{agnihotri2024cospgd} &2.87 & 57.74  &  58.50  & 47.05 \\
& RP-PGD~\cite{zhang2025rp} & \textbf{0.82} & 63.32 & 63.82 &  46.67 \\
& DAG~\cite{xie2017adversarial} & 63.36 & 66.28  & 66.06  & 39.10 \\
& NI-FGSM~\cite{lin2019nesterov} & 10.22& 33.50   & 34.12  & 34.41 \\
& DI-FGSM~\cite{xie2019improving} &7.21 &  29.00  & 30.58 & 39.24 \\
& TI-FGSM~\cite{dong2019evading} &22.23 & 64.64  &  64.95 &  37.29 \\

& FSPGD(Ours) &  2.99 & \textbf{12.48} & \textbf{13.54} & \textbf{21.30} \\
\hline
\multirow{9}*{DV3Res101}& PGD~\cite{mkadry2017towards} & 9.75 & 59.36 & 55.54 & 47.48 \\
& SegPGD~\cite{gu2022segpgd} & 7.18 &54.47  & 53.96 & 46.53 \\
& CosPGD~\cite{agnihotri2024cospgd} & 2.73 &58.83  & 58.54 & 47.25 \\
& RP-PGD~\cite{zhang2025rp} & \textbf{0.68} & 63.71 & 64.53 & 46.91 \\
& DAG~\cite{xie2017adversarial} & 67.55 & 67.09 & 67.58 & 39.48 \\
& NI-FGSM~\cite{lin2019nesterov} &  9.49  &36.41 & 34.75 & 35.62 \\
& DI-FGSM~\cite{xie2019improving} & 7.64  &34.87 & 34.11 &  40.99 \\
& TI-FGSM~\cite{dong2019evading} & 27.16  & 65.79 & 65.13 & 37.98 \\
& FSPGD(Ours) &3.28 & \textbf{11.42} & \textbf{13.45} & \textbf{21.49} \\
\bottomrule
\end{tabular}
\label{table1}
\end{table}


\begin{table}[htbp]
\caption{Attack performance comparison on Cityscapes in terms of mIoU. Lower mIoU means better performance and bold numbers denote the best mIoU values for each experimental setup}
\footnotesize
\begin{tabular}{c|c|ccccc}
\toprule
& & \multicolumn{5}{c}{Target Models (mIoU$\downarrow$)}\\
\hline
\multirow{2}*{Source} & Attack & Source & PSP & DV3 & Segformer & Maskformer \\
\multirow{2}*{Models}&  Method & Model & Res101 & Res101 & MiT-B0 & Swin-S \\
\cmidrule{2-7}
& Clean Images & 64.62 / 65.90 &65.65 &67.16 & 60.58 & 68.24\\
\hline
\multirow{8}*{PSP} & PGD~\cite{mkadry2017towards}& 1.83  & 18.80 & 19.35 & 48.92 & 59.66 \\
\multirow{8}*{Res50}& SegPGD~\cite{gu2022segpgd}& 1.38  & 18.26 & 19.34 & 49.83 & 60.41 \\
& CosPGD~\cite{agnihotri2024cospgd} & 0.07  & 24.90 & 26.65 & 50.31 & 60.53 \\
& RP-PGD~\cite{zhang2025rp} & \textbf{0.05} & 25.85 & 27.62 & 51.25 & 58.70 \\
& DAG~\cite{xie2017adversarial} &23.52 & 36.76 & 33.47 & 50.24 & 60.64 \\
& NI-FGSM~\cite{lin2019nesterov}&  1.62  & 15.07 & 17.07 & 44.43 & 50.09 \\
& DI-FGSM~\cite{xie2019improving}&  1.92  & 17.60 & 21.57 & 52.12 & 54.81 \\
& TI-FGSM~\cite{dong2019evading} &1.64  & 28.39 & 34.07 & 51.91 & 58.70 \\
& FSPGD (Ours) &0.93  &\textbf{5.12}  & \textbf{3.29}  & \textbf{41.30} & \textbf{47.30} \\
\hline
\multirow{8}*{DV3} & PGD~\cite{mkadry2017towards}&  2.00  & 22.19 & 22.06 & 50.28 & 60.64 \\
\multirow{8}*{Res50}& SegPGD~\cite{gu2022segpgd} & 0.96  & 22.20 & 22.51 & 50.59 & 60.24 \\
& CosPGD~\cite{agnihotri2024cospgd} & \textbf{0.01}  & 25.43 & 27.22 & 50.48 & 59.86 \\
& RP-PGD~\cite{zhang2025rp} & 0.04 & 26.09 & 26.83 & 50.70 & 61.23 \\
& DAG~\cite{xie2017adversarial} & 36.54 & 39.01 & 35.98 & 51.59 & 60.19 \\
& NI-FGSM~\cite{lin2019nesterov} &1.55  & 16.65 & 18.26 & 45.76 & 49.89 \\
& DI-FGSM~\cite{xie2019improving} & 2.32  & 19.87 & 23.61 & 52.63 & 55.32 \\
& TI-FGSM~\cite{dong2019evading} &1.48  & 31.93 & 35.45 & 52.77 & 59.56 \\
& FSPGD (Ours) & 1.27  & \textbf{6.09}  & \textbf{3.78 } & \textbf{40.74} & \textbf{47.30} \\

\hline
\hline
\multirow{3}*{Source} & Attack & Source & PSP & DV3 & Segformer & Maskformer \\
\multirow{3}*{Models}& Method & Model & Res50 & Res50 & MiT-B0 & Swin-S \\
\cmidrule{2-7}
& Clean Images & 65.65 / 67.16 & 64.62 & 65.90 & 60.58 & 68.24\\
\hline
\multirow{8}*{PSP} & PGD~\cite{mkadry2017towards} &1.80  & 9.71  & 12.80 & 48.83 & 59.57 \\
\multirow{8}*{Res101} & SegPGD~\cite{gu2022segpgd}&0.90  & 10.64 & 12.85 & 60.41 & 59.48 \\
& CosPGD~\cite{agnihotri2024cospgd} & \textbf{0.02}  & 14.02 & 16.41 & 50.75 & 61.01 \\
& RP-PGD~\cite{zhang2025rp} & \textbf{0.02}  & 14.34 & 17.06 & 50.74 & 61.11 \\
& DAG~\cite{xie2017adversarial}  & 35.74 & 23.56 & 33.92 & 51.65 & 61.42 \\
& NI-FGSM~\cite{lin2019nesterov} &1.65  & 8.35  & 10.01 & 42.51 & 46.90 \\
& DI-FGSM~\cite{xie2019improving} &2.18  & 16.94 & 19.39 & 50.57 & 53.31 \\
& TI-FGSM~\cite{dong2019evading} & 1.73  & 25.15 & 29.84 & 50.95 & 57.30 \\
& FSPGD (Ours) & 2.29  & \textbf{5.96}  & \textbf{7.42}  & \textbf{36.63} & \textbf{36.91} \\
\hline
\multirow{8}*{DV3} & PGD~\cite{mkadry2017towards} &1.74  & 15.20 & 16.54 & 49.92 & 60.50 \\
\multirow{8}*{Res101}& SegPGD~\cite{gu2022segpgd} &0.63  & 17.29 & 18.04 & 50.18 & 60.11 \\
& CosPGD~\cite{agnihotri2024cospgd} &\textbf{0.01}  & 18.83 & 19.60 & 50.44 & 59.91 \\
& RP-PGD~\cite{zhang2025rp} & 0.03 & 18.31 & 19.92 & 50.89 & 61.66 \\
& DAG~\cite{xie2017adversarial} &36.68 & 26.70 & 36.68 & 52.25 & 61.29 \\
& NI-FGSM~\cite{lin2019nesterov} & 1.94  & 14.15 & 14.91 & 44.14 & 48.74 \\
& DI-FGSM~\cite{xie2019improving} & 3.99  & 22.41 & 23.87 & 50.91 & 53.96 \\
& TI-FGSM~\cite{dong2019evading} & 2.63  & 29.58 & 32.17 & 51.90 & 57.19 \\
& FSPGD (Ours) & 2.03  & \textbf{2.48}  &\textbf{ 3.25 } & \textbf{39.82 }& \textbf{47.04}\\
\bottomrule
\end{tabular}
\label{table2}
\end{table}

\section{Experiment}
\label{sec4:experiment}

\subsection{Experimental Setup}
\label{sec4.1:Experimental Setup}

\noindent\textbf{Datasets.}
We evaluate on two widely used semantic segmentation benchmarks: PASCAL VOC 2012~\cite{pascal-voc-2012} and Cityscapes~\cite{cordts2016cityscapes}. 
VOC contains 20 foreground classes and one background class, with 1{,}464 training and 1{,}499 validation images. 
Following standard practice~\cite{hariharan2015hypercolumns}, the training split is expanded to 10{,}582 images. 
Cityscapes provides high quality pixel annotations over 19 classes with 2{,}975 training and 500 validation images. 
All results are reported on the validation splits. 
Unless otherwise noted, we adopt the standard evaluation resolution and preprocessing of each model as provided in its public implementation so that the comparison reflects the behavior of the attack rather than changes of inference protocol.

\noindent\textbf{Models and transfer protocol.}
We consider both convolutional and transformer architectures. 
For convolutional networks we use PSPNet with ResNet 50 and ResNet 101 encoders and DeepLabv3 with ResNet 50 and ResNet 101 encoders~\cite{zhao2017pyramid, chen2017rethinking, he2016deep}. 
For additional diversity on the target side we include FCN with a VGG16 encoder~\cite{long2015fully}. 
For transformers we use SegFormer with MiT B0~\cite{xie2021segformer} and Mask2Former with Swin Small~\cite{cheng2022masked, liu2021swin}. 
We alternate source and target roles to measure transferability across many pairs. 
To enforce a genuine black box transfer setting, the source and target models never share the same encoder. 
We also include transfers across architectural families, for example convolutional to transformer and transformer to convolutional, to emulate realistic deployment where the target architecture is unknown.

\noindent\textbf{Feature layer for FSPGD.}
FSPGD perturbs intermediate representations extracted from a fixed layer of the surrogate. 
Unless otherwise specified, we use the layer choices identified by our ablation studies as yielding the best average transferability. 
For ResNet 50 we use the second block in the Conv3\_x stage. 
For ResNet 101 we use the tenth block in the Conv3\_x stage. 
For MiT B0 we use the first layer of Transformer Block 1. 
For Swin Small we use the first block of Stage 2. 
These defaults are used throughout the main tables, while full layer sweeps are reported in the ablation section.

\noindent\textbf{Threat model and attack parameters.}
All attacks follow the $\ell_\infty$ threat model with maximum perturbation $\epsilon=8/255$. 
We use a random start drawn uniformly from $[-\epsilon,\epsilon]$ and update for $T=20$ iterations with step size $\alpha=2/255$, projecting after each step onto $\mathcal{B}_\infty(x,\epsilon)$ and clamping to the valid image domain. FSPGD uses the dual loss with the iteration dependent weight $\lambda_t=t/T$ and the binary selection mask threshold $\tau=\cos(\pi/3)$ in Eq.~\ref{eq7}. 
Unless noted, these settings are kept fixed for all source target pairs to isolate the effect of the attack objective.
\begin{table}[t]
\caption{Attack performance comparison on Cityscapes in terms of mIoU. Lower mIoU means better performance and bold numbers denote the best mIoU values for each experimental setup}
\setlength{\tabcolsep}{6pt}
\footnotesize

\begin{tabular}{c |c |c c c c c c}
\toprule
& & \multicolumn{6}{c}{Target Models (mIoU$\downarrow$)}\\
\hline
\multirow{2}*{Source} & Attack & Source & PSP & DV3 & PSP & DV3 & Mask2Former \\
\multirow{2}*{Models}& Method & Model & Res50 & Res50 &  Res101 & Res101 & Swin-S \\
\cmidrule{2-8}
& Clean Images & 60.58 &64.62 & 65.65 & 65.90 & 67.16 & 68.24 \\
\hline
\multirow{8}*{SegFormer} & PGD~\cite{mkadry2017towards} &1.06& 29.94& 36.07& 31.99& 38.25& 48.43\\
\multirow{8}*{MiT-B0}& SegPGD~\cite{gu2022segpgd} &0.38& 28.45& 34.56& 29.28& 36.38& 49.54\\
& CosPGD~\cite{agnihotri2024cospgd} & \textbf{0.00} & 29.98& 35.92& 32.19& 37.72& 51.51\\
& RP-PGD~\cite{zhang2025rp} & 0.03 & 32.19 & 32.45 & 32.45 & 39.42 & 53.11 \\
& DAG~\cite{xie2017adversarial} &50.92& 20.84& 33.73& 32.71& 28.77& 55.21\\
& NI-FGSM~\cite{lin2019nesterov} &2.06& 30.27& 37.63& 30.95& 38.24& 43.75\\
& DI-FGSM~\cite{xie2019improving} & 9.13& 41.92& 45.85& 43.10& 48.06& 46.78\\
& TI-FGSM~\cite{dong2019evading} & 7.66& 50.60& 52.77& 52.25& 55.88& 55.59\\

& FSPGD (Ours)  &1.33& \textbf{10.09} & \textbf{14.57} & \textbf{21.16} & \textbf{22.06} & \textbf{39.92}
\\
\hline
\hline
\multirow{2}*{Source} & Attack & Source&PSP & DV3&  PSP & DV3 & SegFormer\\
\multirow{2}*{Models}& Method & Model & Res50 & Res50 &  Res101 & Res101 &MiT-B0 \\
\cmidrule{2-8}
& Clean Images &  68.24 & 64.62 & 65.65& 65.90 & 67.16 & 60.58 \\
\hline
\multirow{8}*{Mask2Former} & PGD~\cite{mkadry2017towards} &0.45 & 39.41 & 45.25& 42.15& 48.35& 49.30\\
\multirow{8}*{Swin-S} & SegPGD~\cite{gu2022segpgd} &0.30& 39.97& 45.07& 42.29& 48.96& 49.40\\
& CosPGD~\cite{agnihotri2024cospgd} &\textbf{0.17} & 39.56& 45.23& 42.36& 47.43& 49.37\\
& RP-PGD~\cite{zhang2025rp} & 0.32 & 40.01 & 42.89 & 45.52 & 490.6 & 49.99\\
& DAG~\cite{xie2017adversarial} & 65.59& 30.69& 42.06& 32.76& 39.42& 54.23\\
& NI-FGSM~\cite{lin2019nesterov} & 0.17& 42.76& 49.41& 45.06& 50.00& 45.87\\
& DI-FGSM~\cite{xie2019improving} &3.53& 50.34& 53.67& 53.16& 56.59& 50.85\\
& TI-FGSM~\cite{dong2019evading} &  0.85& 56.81& 59.74& 59.95& 62.69& 59.74\\
& FSPGD (Ours)  &2.20  & \textbf{15.57} &  \textbf{18.00}&\textbf{24.29} & \textbf{25.96} & \textbf{36.87 }\\
\bottomrule
\end{tabular}
\label{table3}
\end{table}
\noindent\textbf{Baselines.}
We compare against representative transfer attacks widely used in prior work, including PGD~\cite{mkadry2017towards}, NI-FGSM~\cite{lin2019nesterov}, DI-FGSM~\cite{xie2019improving}, and TI-FGSM~\cite{dong2019evading}. 
We further include segmentation specific baselines SegPGD~\cite{gu2022segpgd} and CosPGD~\cite{agnihotri2024cospgd}, as well as DAG~\cite{xie2017adversarial}, an early dense prediction attack that serves as a reference point for segmentation and detection. Given recent advances in segmentation attacks, we additionally evaluate RP-PGD~\cite{zhang2025rp}. Across both datasets and across the tested source target pairs, FSPGD achieves lower adversarial mIoU on average than RP-PGD under the same $\epsilon$, $\alpha$, and $T$.

\noindent\textbf{Metrics and reporting.}
We report mean Intersection over Union in percent. 
For transfer attacks, lower mIoU on adversarial examples indicates stronger attack performance on the target model. 
For adversarial training experiments, where models are trained on perturbations from a given attack and then evaluated against unseen attacks, higher mIoU indicates better robustness. 
Unless stated otherwise, each table reports clean mIoU and adversarial mIoU for the target, and we also discuss the difference between the two as an aggregate measure of transfer strength. 
We follow the same metric protocol on both datasets to allow direct comparison across architectures and training recipes.

\subsection{Quantitative Results}
\label{sec4.2:Expresult}

We report comprehensive transfer-based black-box attack results across two datasets and multiple architecture pairings. On PASCAL VOC (Table~\ref{table1}), where both sources and targets are convolutional, FSPGD consistently achieves the lowest adversarial mIoU among all competitors while preserving strong white-box degradation on the surrogate. This establishes a solid baseline that isolates the effect of our feature-space objective under a conventional CNN$\rightarrow$CNN setting.

On Cityscapes, we evaluate two complementary transfer regimes. First, Table~\ref{table2} considers CNN sources (PSPNet–ResNet50 and DeepLabV3–ResNet50) with both CNN and Transformer targets (PSPNet–ResNet101, DeepLabV3–ResNet101, SegFormer–MiT-B0, Mask2Former–Swin-S).  Across all target models, FSPGD yields markedly lower mIoU than PGD~\cite{mkadry2017towards}, NI-FGSM~\cite{lin2019nesterov}, DI-FGSM~\cite{xie2019improving}, TI-FGSM~\cite{dong2019evading} and segmentation-specific baselines such as SegPGD~\cite{gu2022segpgd}, CosPGD~\cite{agnihotri2024cospgd} and RP-PGD~\cite{zhang2025rp}, indicating that perturbing intermediate representations with the proposed dual loss amplifies cross-model drift. The gains are most pronounced for cross-architecture transfers (CNN $\rightarrow$ Transformer), where methods tailored to output-logit objectives frequently underperform: FSPGD reduces mIoU on SegFormer and Mask2Former targets by large margins compared with momentum/diversity/translation baselines, confirming that representation-level attacks address failure modes not captured by output-level losses.

Second, Table~\ref{table3} reverses the direction and uses SegFormer–MiT-B0 and Mask2Former–Swin-S while evaluating against both CNN and Transformer targets. This setting is particularly informative because source–target inductive biases differ substantially. FSPGD again attains the best or second-best mIoU in almost every configuration and demonstrates clear advantages on CNN targets when crafted from Transformer surrogates, highlighting that our internal feature diversification alleviates the over-consistency that hampers transfer in dense prediction. Together, Tables~\ref{table2} and ~\ref{table3} show that FSPGD is robust to architecture mismatch in both directions (CNN$\leftrightarrow$Transformer), whereas many prior attacks degrade sharply once the source and target families diverge.

\begin{figure*}[t]
\centering
\includegraphics[width=1\linewidth]{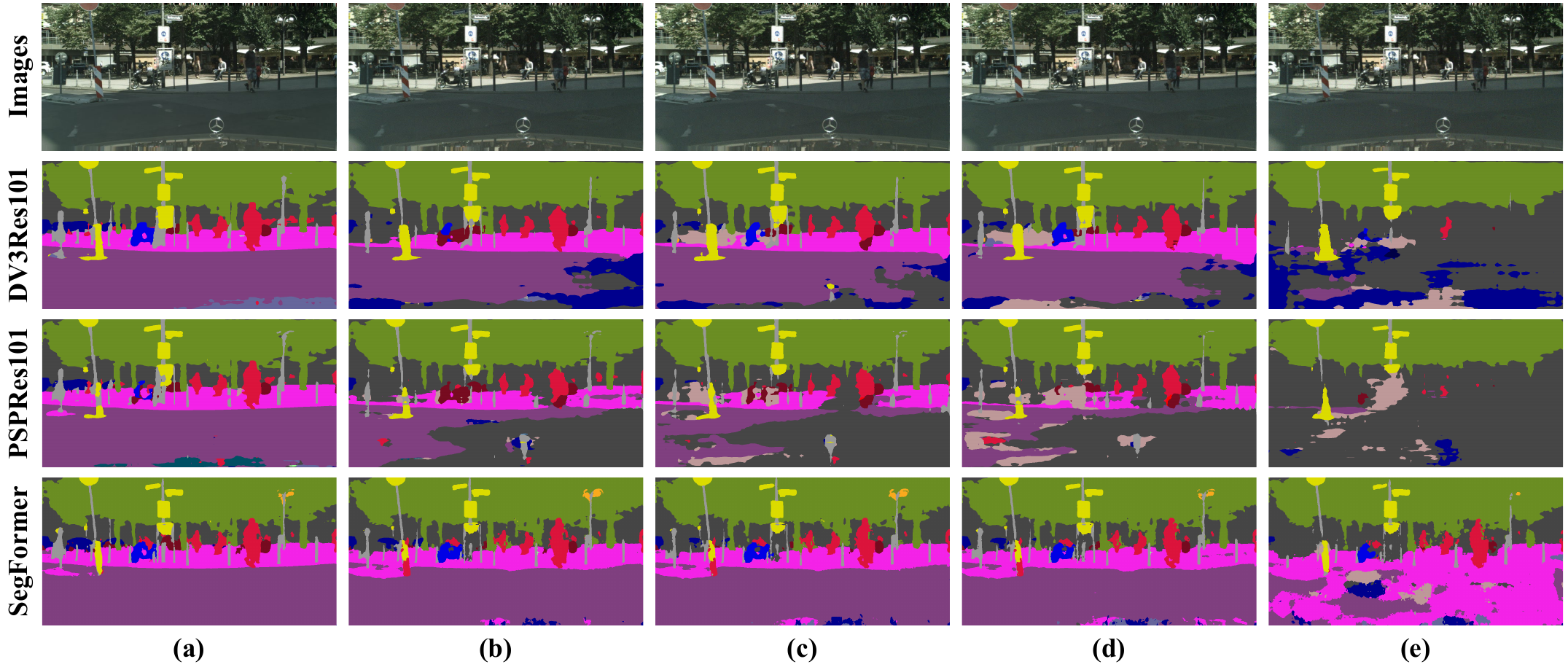}
\caption{Visualization of clean image, attacked images, and output predictions on Cityscapes.  Deeplabv3-Res101 is used as the source model and Deeplabv3-Res101 (second row), PSPNet-Res101 (third row), and SegFormer (fourth row) are used as target models. (a) Clean image, (b) PGD, (c) SegPGD, (d) CosPGD, (e) FSPGD (Ours).}
\label{fig:fig4}
\end{figure*}

\subsection{Qualitative Results}
\label{sec4.3:qual}
Figure~\ref{fig:fig4} illustrates qualitative comparisons on Cityscapes. Using DeepLabV3–ResNet50 as the surrogate, we visualize clean images, adversarial images, and target predictions for PSPNet/DeepLabV3 with ResNet-50/101 and for Transformer targets. Conventional attacks often preserve large portions of the clean segmentation, reflecting limited cross-model drift and, consequently, weak transfer. In contrast, FSPGD systematically disrupts object masks and boundaries across all targets. The qualitative artifacts are consistent with our design: early-stage internal loss reduces similarity among spatially separated same-class regions, which fragments context and breaks label propagation, while the late-stage external loss drives clean–adversarial feature discrepancy, preventing the targets from mapping adversarial features back to clean decisions. The visual patterns align with the numerical gaps observed in Tables~\ref{table1},~\ref{table2}, and ~\ref{table3}, and make clear that the improvements are not confined to a specific architecture or scene category.

\begin{table}[t]
\caption{Robustness comparison on Pascal VOC 2012 in terms of mIoU. Higher mIoU indicates stronger robustness. Bold numbers denote the best performance in each attack setting.}
\begin{tabular}{c|c|ccccc}
\toprule
\multicolumn{1}{c}{}& & \multicolumn{5}{c}{Target Models( mIoU↑)} \\
\hline
\multirow{3}{*}{Source} & Attack methods & \multicolumn{5}{c}{PSPRes50} \\
\cmidrule{2-7}
Models & \diagbox{Attack}{Training} & PGD & SegPGD & CosPGD &RP-PGD  & FSPGD\\
\hline
\multirow{4}{*}{PSP} & PGD~\cite{mkadry2017towards} & 39.68 & 41.86 & 40.08 & 37.83& \textbf{55.35}\\
\multirow{4}{*}{Res101} & SegPGD~\cite{gu2022segpgd} & 39.10 & 41.73 & 39.73 & 37.60 &\textbf{55.33} \\
 & CosPGD~\cite{agnihotri2024cospgd} & 39.65 & 41.95 & 39.99 & 37.99& \textbf{55.24}  \\
 & RP-PGD~\cite{zhang2025rp} & 39.90 & 41.82 & 40.34 & 37.44 &\textbf{55.40}  \\
 & FSPGD (Ours) & 23.89 & 27.96 & 25.07 &21.89& \textbf{53.06} \\
 
\hline

\multirow{4}{*}{DV3} &  PGD~\cite{mkadry2017towards} & 39.89 & 41.97 & 40.31&38.63 & \textbf{55.39}  \\
\multirow{4}{*}{Res101} & SegPGD~\cite{gu2022segpgd} & 39.87 & 41.85 & 40.29 & 38.15& \textbf{55.30}  \\
& CosPGD~\cite{agnihotri2024cospgd}& 39.90 & 41.74 & 40.25 &38.23& \textbf{55.31} \\
& RP-PGD~\cite{zhang2025rp}& 39.15 & 41.64 & 39.63 &38.09& \textbf{55.26} \\
& FSPGD (Ours) & 22.83 & 27.34 & 24.23 &18.85&\textbf{53.13}\\
\hline
\hline

\multicolumn{1}{c}{}& &\multicolumn{5}{c}{DV3Res50} \\
\hline

 & \diagbox{Attack}{Training} & PGD & SegPGD & CosPGD &RP-PGD  & FSPGD\\

\hline
\multirow{4}{*}{PSP} & PGD~\cite{mkadry2017towards} & 33.86 & 38.12 & 37.00 & 36.33 & \textbf{55.06} \\
\multirow{4}{*}{Res101} & SegPGD~\cite{gu2022segpgd} &  33.53 & 37.89 & 36.72 &36.06& \textbf{55.19} \\
 & CosPGD~\cite{agnihotri2024cospgd} & 33.84 & 38.35 & 37.42& 36.20 & \textbf{55.32} \\
 & RP-PGD~\cite{zhang2025rp} & 33.97 & 38.48 & 37.27& 36.02 & \textbf{55.47} \\
 & FSPGD (Ours)  & 17.71 & 25.88 & 22.97& 20.06 & \textbf{53.59} \\
\hline

\multirow{4}{*}{DV3} &  PGD~\cite{mkadry2017towards} & 34.36 & 38.32 & 37.46 &36.74 &\textbf{55.05} \\
\multirow{4}{*}{Res101} & SegPGD~\cite{gu2022segpgd}& 33.88 & 37.99 & 36.86 &36.31& \textbf{55.02} \\
&  CosPGD~\cite{agnihotri2024cospgd} & 34.08 & 38.69 & 37.55 &36.41& \textbf{55.25} \\
& RP-PGD~\cite{zhang2025rp}&  33.32 & 38.16 & 36.80 &36.27& \textbf{55.31} \\
& FSPGD (Ours) & 16.90 & 25.38 & 22.36 &21.12& \textbf{53.65}\\
\bottomrule
 
\end{tabular}
\label{table4}
\end{table}

\begin{table}[htbp]
\caption{Comparison of mIoU scores on SAT and DDC-AT defense settings using various attack methods.}
\setlength{\tabcolsep}{5pt}
\footnotesize
\begin{tabular}{c|c|cccc|cccc}
\toprule
\multirow{3}{*}{Source} & \multirow{3}{*}{Attack Method} & \multicolumn{4}{c|}{SAT} & \multicolumn{4}{c}{DDC-AT} \\
\cmidrule{3-10}
\multirow{3}{*}{Models}& & PSP & DV3 & PSP & DV3 & PSP & DV3 & PSP & DV3 \\
& & Res50 & Res50 & Res101 & Res101 & Res50 & Res50 & Res101 & Res101 \\
\cmidrule{2-10}
 & Clean Images & 73.53 & 69.69 &76.34 & 75.52 & 74.00 & 74.25 & 70.38 & 76.81 \\
\hline
\multirow{4}{*}{PSP}& PGD~\cite{mkadry2017towards} & 69.25 & 62.27 & 72.37 & 71.38 & 70.89 & 69.01 & 68.20 & 72.82 \\
\multirow{4}{*}{Res101}& SegPGD~\cite{gu2022segpgd}& 69.34 & \textbf{62.10} & 72.13 & 71.44 & 70.74 & 69.19 & 68.33 & 72.88 \\
& CosPGD~\cite{agnihotri2024cospgd}  &69.42 & 62.43 & 72.13 & 71.74 & 70.69 & 69.48 & 68.27 & 72.95 \\
& RP-PGD~\cite{zhang2025rp} & 69.25 & 62.27 & 72.37 & 71.38 & 70.89 & 69.01 & 68.20 & 72.82 \\
& FSPGD (Ours)  &\textbf{ 60.37 }& 67.02 & \textbf{69.28 }& \textbf{69.80 }& \textbf{69.63 }& \textbf{68.63} & \textbf{67.76} & \textbf{70.30} \\

\hline
\multirow{5}{*}{DVRes101}& PGD~\cite{mkadry2017towards} & 69.21 & 62.14 &71.99&71.46& 71.02 & 69.18&68.30& 72.72\\
& SegPGD~\cite{gu2022segpgd} & 69.08 & \textbf{62.01}&71.72& 71.30& 70.88 & 68.86&\textbf{68.26}& 72.62\\
& CosPGD~\cite{agnihotri2024cospgd} & 69.17 & 62.08& 71.77& 71.56 & 70.90 & 69.35&68.27& 72.61 \\
& RP-PGD~\cite{zhang2025rp} & 69.43  & 62.36 & 72.20 & 71.06& 70.82 & 69.45& 68.31& 72.69\\
& FSPGD (Ours)  & \textbf{57.77} & 62.06 & \textbf{51.86}& \textbf{68.98}& \textbf{66.13} & \textbf{66.56}&68.62& \textbf{66.27}\\
\bottomrule
\end{tabular}
\label{table5}
\end{table}

\subsection{Adversarial Training (Robustness)}
\label{sec4.4:advtrain}
We study robustness from two complementary perspectives. 
First, we ask whether adversarial examples generated by each attack serve as effective training signals when used for adversarial training. 
Second, we examine how well each attack succeeds against models that have already been trained with standard defenses.

\noindent\textbf{Training with attack-specific examples.}
We further assess whether adversarial examples generated by different attacks improve robustness under adversarial training. Following standard practice, each model is trained with one attack and then evaluated against multiple unseen attacks under identical budgets. As summarized in Table~\ref{table4}, training with FSPGD adversarial examples yields competitive or superior mIoU under strong evaluation attacks on the PSPNet-ResNet50 and DeepLabV3-ResNet50 targets, supporting the intuition that disrupting external clean-adversarial alignment and intraclass consistency produces harder, more diverse training signals than logit-only objectives. This result complements our transfer findings and provides an orthogonal robustness benefit.

\noindent\textbf{Attacking defended models.}
We further evaluate the effectiveness of FSPGD against models trained with two representative defense strategies, SAT and DDC-AT~\cite{xu2021dynamic}, using identical budgets for all attacks. 
The results, summarized in Table~\ref{table5}, highlight three important observations. 
First, under SAT, FSPGD substantially reduces the performance of defended models: for instance, PSPNet–ResNet50 drops from 73.53 on clean inputs to 57.77 under attack, a reduction of nearly 16 percentage points. This represents the strongest degradation among compared methods and demonstrates that FSPGD adversarial examples are sufficiently diverse to bypass the robustness conferred by standard adversarial training. 
Second, on DeepLabV3–ResNet50, FSPGD achieves an adversarial mIoU of 62.06, which is comparable to or stronger than specialized segmentation attacks such as SegPGD, confirming that our feature-space perturbations are competitive even against methods tailored for this domain. 
Third, under the more advanced DDC-AT defense, FSPGD consistently attains the lowest adversarial mIoU across both PSPNet–ResNet50 (66.13) and DeepLabV3–ResNet50 (66.56), outperforming PGD, SegPGD, and CosPGD. 
Together, these results demonstrate that FSPGD retains its effectiveness even when the target models have been adversarially trained, underscoring the resilience of feature-space perturbations in breaking defense-aware robustness.

\noindent\textbf{Interpretation.}
Robustness gains with FSPGD adversarial training in Table~\ref{table4} and the attack effectiveness against defended models in Table~\ref{table5} are consistent with the design of our objective. 
By enforcing representational drift between clean and adversarial features and by reducing redundant similarity among spatially separated same-class regions, FSPGD generates perturbations that are both diverse and persistent in the intermediate representation space. 
Such perturbations make defended models less able to project adversarial features back to clean decisions, and when used during training they expose richer failure modes that translate into improved robustness against unseen attacks.

\subsection{Ablation Studies}
\label{sec4.5:Ablation}
We conduct ablation studies to validate the main design choices of FSPGD. The analysis covers three aspects: the threshold $\tau$ for binarizing the similarity mask $M$, the contribution and scheduling of the two loss components $L_{ex}$ and $L_{in}$, and the choice of intermediate feature layers.

\begin{figure}[t]
\centering
\includegraphics[width=0.5\linewidth]{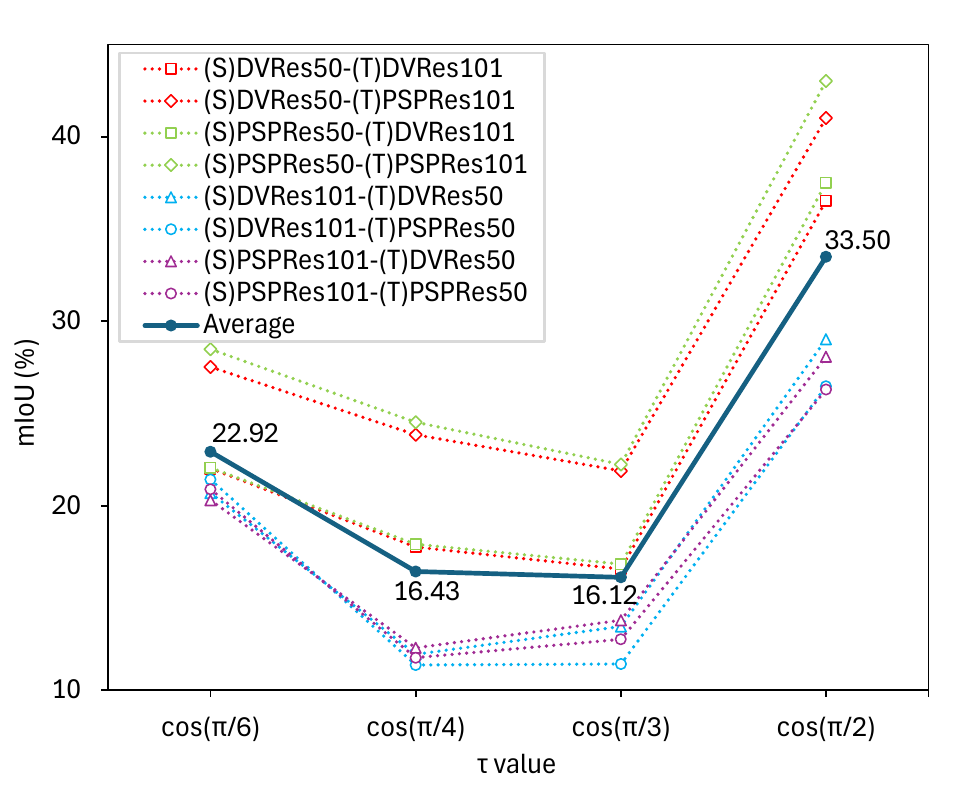}
\caption{mIoU performance across different $\tau$ values. (S) and (T) indicate the source and target models, respectively.}
\label{fig:fig6}
\end{figure}

\textbf{Effect of threshold $\tau$.}
The mask $M_B$ (Eq.~\ref{eq7}) selects pixel pairs that are similar in the clean feature space. Varying $\tau$ changes how strictly these pairs are selected. As shown in Fig.~\ref{fig:fig6}, transferability is maximized when $\tau=\cos(\pi/3)$. Smaller thresholds introduce noisy pairs, while larger thresholds leave too few, both reducing effectiveness. Across all settings FSPGD consistently outperforms logit-level baselines, confirming robustness to moderate variations of $\tau$.

\begin{figure}[t]
\centering
\includegraphics[width=0.5\linewidth]{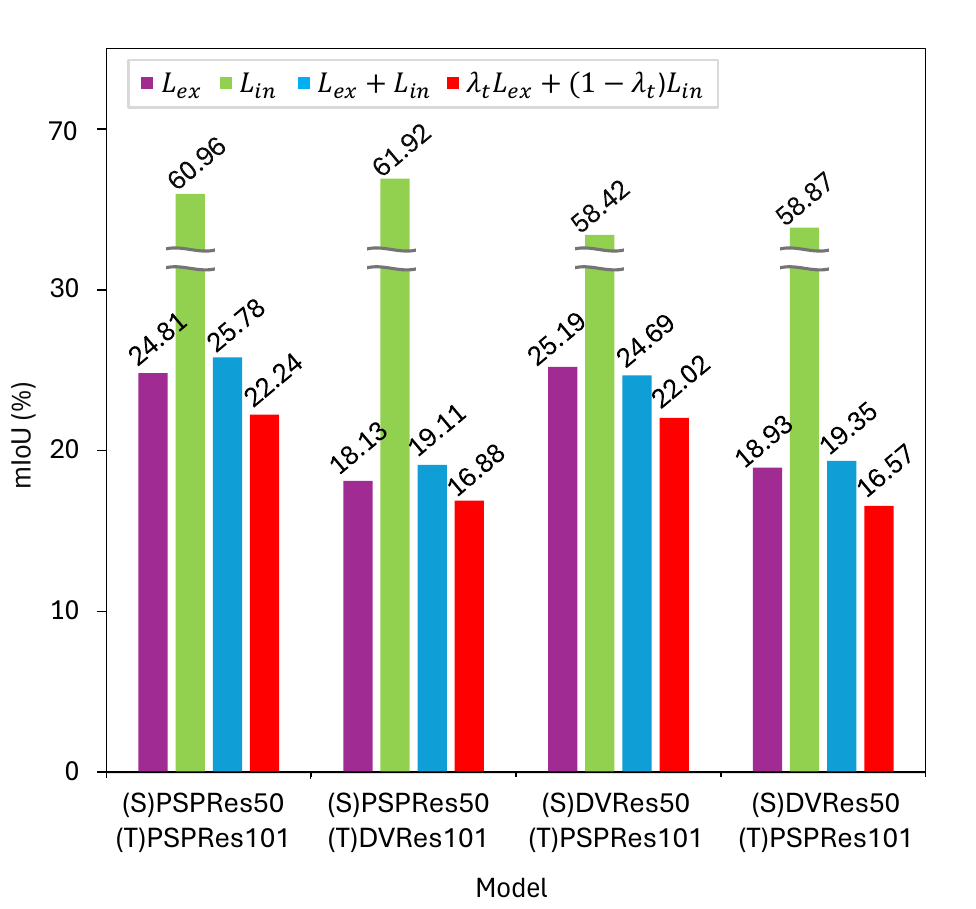}
\caption{mIoU performance across different loss terms. (S) and (T) indicate the source and target models, respectively.}
\label{fig:fig7}
\end{figure}

\textbf{Role of $L_{ex}$ and $L_{in}$.}
We compare four variants: using only $L_{ex}$, only $L_{in}$, a fixed equal-weight combination, and the dynamic schedule $\lambda_t=t/T$ (Fig.~\ref{fig:fig7}). $L_{ex}$ alone is effective, whereas $L_{in}$ alone is weak. The fixed combination does not surpass $L_{ex}$. In contrast, the staged schedule significantly improves transferability by letting $L_{in}$ diversify same-class features in early steps, followed by $L_{ex}$ pushing them away from clean features in later steps. This confirms that $L_{in}$ is complementary and most effective when used with scheduling.

\begin{figure}[htbp]
\centering
\includegraphics[width=1\linewidth]{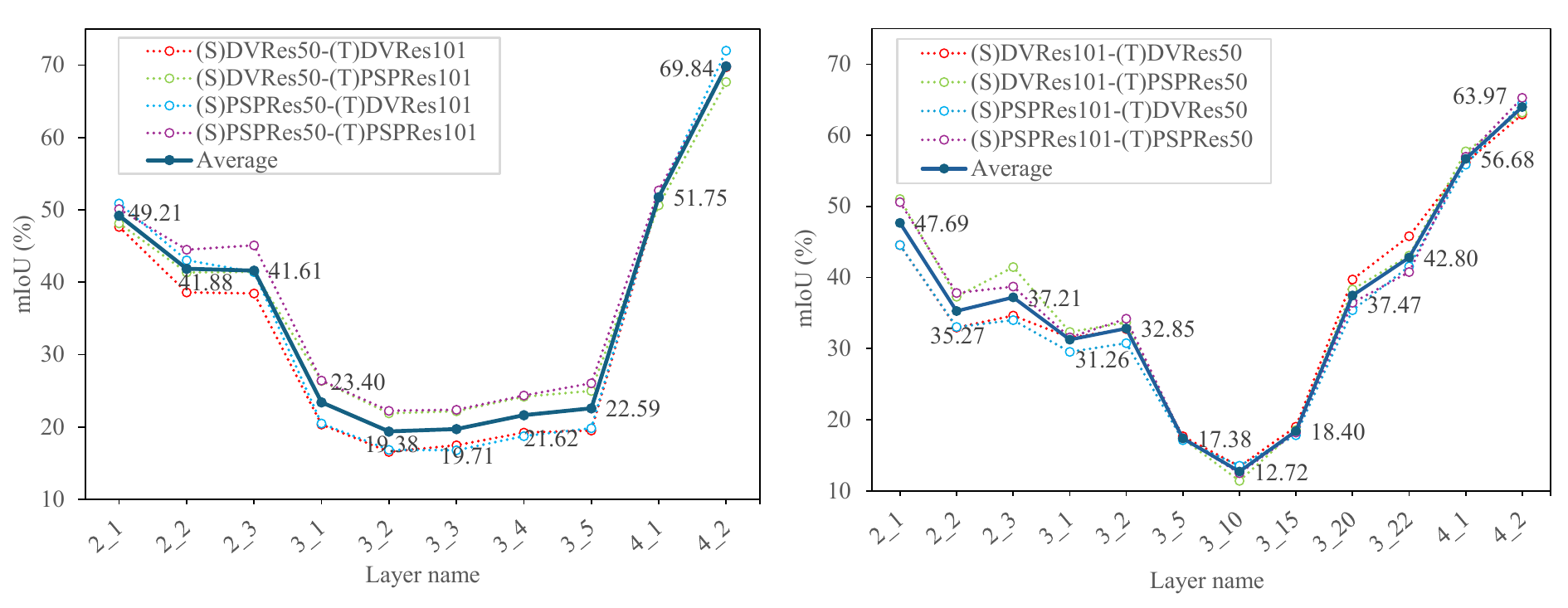}
\caption{mIoU performance across different locations of intermediate layers. (S) and (T) indicate the source and target models, respectively.}
\label{fig:fig8}
\end{figure}

\textbf{Choice of feature layer.}
We test different layers in both CNN and transformer backbones (Fig.~\ref{fig:fig8}). Middle encoder layers yield the best transferability, while early layers capture only low-level textures and late layers overfit to model-specific logits. The results highlight that semantically meaningful yet generalizable features in mid-level layers are the most effective attack space.

In summary, the ablations confirm that (i) $\tau=\cos(\pi/3)$ balances meaningful selection and noise, (ii) $L_{in}$ and $L_{ex}$ are complementary and require scheduling, and (iii) perturbing middle-layer features achieves the best transferability. These results explain the consistent improvements of FSPGD across benchmarks.

\section{Discussion}
\label{sec:discussion}

The experimental results and ablation studies highlight two central insights. 
First, effective transferability in semantic segmentation cannot be achieved by perturbing output logits alone. 
Our analysis of intra-class consistency shows that conventional methods fail to diversify features of spatially separated instances, which limits their ability to generalize to unseen models. 
By explicitly disrupting this over-consistency, FSPGD generates perturbations that remain effective even under significant architectural shifts, such as CNN-to-Transformer transfer.

Second, the staged use of the dual-loss design is critical. 
Although the internal loss $L_{in}$ is weak in isolation, its role in early iterations is to diversify the feature space, thereby enhancing the impact of the external loss $L_{ex}$ in later updates. 
This synergy explains the superior transferability of FSPGD compared with conventional segmentation attacks and emphasizes the value of intermediate-layer perturbations as an overlooked dimension in dense prediction tasks. 

Finally, while FSPGD introduces user-defined parameters ($\tau$ and $\lambda_t$), our sensitivity analyses confirm that the method is stable across a wide range of values. 
This indicates that FSPGD does not rely on fragile hyperparameter tuning, although adaptive parameter selection remains a promising avenue for future research. 
Overall, these findings suggest that robust transferability in semantic segmentation requires methods that explicitly engage with feature-space representations, and FSPGD provides one practical approach in this direction.

\section{Conclusion}
\label{sec:conclusion}

In this paper, we identify key limitations in existing segmentation attack methods and conduct an in-depth analysis of the underlying causes. Based on these observations, we develop and introduce a novel segmentation attack method, called Feature Similarity Projected Gradient Descent (FSPGD), specifically designed to enhance both attack performance and transferability. The proposed FSPGD method demonstrates notable improvements over conventional methods, not only in terms of attack efficacy but also in transferability across different model architectures. To validate the superiority of FSPGD, we conduct extensive experiments across various source and target models, including PSPNet-ResNet50, PSPNet-ResNet101, DeepLabv3-ResNet50, DeepLabv3-ResNet101, SegFormer MiT-B0 and Mask2Former Swin-S. Future work will aim to further optimize the parameter settings of FSPGD to enhance its robustness and adaptability across various model configurations. Additionally, we plan to explore a more automated approach for parameter optimization, which would allow the method to achieve optimal results efficiently across a diverse set of models, thus broadening its applicability in real-world scenarios.

\backmatter




\section*{Acknowledgements}
This work was supported by the National Research Foundatation of Korea (NRF) grant funded by the Korea government (MSIT) (Grant\#. RS-2025-00517079). This work was supported by ITRC support program supervised by the IITP and funded by MSIT (Grant\#: IITP-2025-RS-2023-00258971).

\section*{Declarations}

\begin{itemize}

\item \textbf{Competing interests} The authors declare that they have no competing interests.






\end{itemize}

\noindent

\bigskip

\bibliography{mainbib}

\end{document}